%% file: MutuallyConflictingActionTimes.tex
\def\year{2020}\relax

\documentclass[letterpaper]{article}
\usepackage{lmodern}
\usepackage{amssymb,amsmath}

\usepackage{aaai20}
\usepackage{times}
\usepackage{helvet}
\usepackage{courier}
\usepackage{graphicx}
\usepackage{algorithm}
\usepackage{algpseudocode}
\usepackage{xcolor}
\usepackage{amsthm}
\usepackage{color}
\usepackage{verbatim}
\usepackage{tikz}
\usepackage{pgfplots}
\usetikzlibrary{arrows.meta,shapes}
\usetikzlibrary{pgfplots.groupplots}
\usepackage{multirow}
\usepackage{enumerate}
\usepackage{booktabs}
\usepackage{nicefrac}
\usepackage{dblfloatfix}

\usepackage[margin=1in]{geometry}
\usepackage{hyperref}
\hypersetup{unicode=true,
            pdftitle={Collision Detection for Agents in Multi-Agent Pathfinding},
            pdfauthor={Thayne T. Walker; Nathan R. Sturtevant},
            pdfborder={0 0 0},
            breaklinks=true}
\usepackage{graphicx,grffile}

  \title{Collision Detection for Agents in Multi-Agent Pathfinding}

    \author{Thayne T. Walker$^1$, Nathan R. Sturtevant$^2$ \\
$^1$ University of Denver, Denver, USA \\
$^2$ University of Alberta, Edmonton, Canada}

\begin{document}
\maketitle
\begin{abstract}
Recent work on the multi-agent pathfinding problem (MAPF) has begun to study agents with motion that is more complex, for example, with non-unit action durations and kinematic constraints. An important aspect of MAPF is collision detection. Many collision detection approaches exist, but often suffer from issues such as high computational cost or causing false negative or false positive detections. In practice, these issues can result in problems that range from inefficiency and annoyance to catastrophic. The main contribution of this technical report is to provide a high-level overview of major categories of collision detection, along with methods of collision detection and anticipatory collision avoidance for agents that are both computationally efficient and highly accurate.
\end{abstract}

\section{Introduction}\label{introduction}

Multi-agent pathfinding (MAPF) is the problem of finding paths for a set of agents from respective start locations to goal locations in a shared space while avoiding conflicts. MAPF has applications in robotics, navigation, games, etc. The problem of detecting collisions between agents (robots, objects, players, etc.) is of central importance for MAPF. There is a fundamental trade-off between accuracy and computation time; more accurate collision detection is often preferred in systems with strict safety requirements (e.g. human transport) while others may trade accuracy for speed when safety is not an issue (e.g. games).

This technical report first provides definitions and background for collision detection, a broad overview and categorization of existing methods and summarizes the advantages and disadvantages of each. Then efficient and exact equations for collision detection for circular and spherical agents with constant velocity and initial velocity with constant acceleration are formally defined. Finally, conic equations for anticipatory collision avoidance for circular and spherical agents are introduced. Many examples are used to illustrate the problem
for 2-dimensional spaces, however, the case of 3-dimensional spaces is directly applicable.

\section{Background}\label{background}

Conflict detection is important for many problems with multiple moving agents. For the purposes of this technical report, agents have spatial locations and a physical shapes such as circles, spheres, polygons or polygonal meshes. Agents may only occupy one location at a time, situated using a \emph{reference point} \cite{li2019large}. In the case of navigation and routing problems for multiple agents, feasible joint solutions cannot be found or verified without proper conflict detection. A conflict represents a simultaneous attempt to access a joint resource. Depending on the target domain a conflict may have different meanings, for example when states have dimensions other than spatial components such as scheduling problems or resource allocation problems (e.g. allocating time slots for classrooms or coordinating memory and cpu allocation for processing jobs) or when abstract states are used such as in dimensionally-reduced spaces. Typically, when considering only temporospatial aspects, conflict detection is referred to as \emph{collision detection}.

Collision detection has been extensively studied in the fields of computational geometry, robotics, and computer graphics. When selecting a method for checking conflicts we need to be cognizant of \emph{type I} and \emph{type II} errors \cite{neyman1933testing}, that is, false positives (reporting a conflict that does not actually occur) and false negatives (not reporting a conflict that actually does occur). A method that exhibits type II errors should never be used because type II errors can lead to infeasible solutions. A method that exhibits type I errors may be used, but may be incomplete or lead to sub-optimal solutions. In this section we provide a brief taxonomy of collision detection techniques for multiple moving agents.

This tech report focuses on solutions for \emph{segmented motion}. Segmented motion is defined as a series of movements (or actions) for agents that have a discrete length. Segmented motion is the natural product of path planning  for agents in discretized spaces such as grids, graphs and robotic latices. A segment of motion is defined by a pair of states: $\langle s_{start}, s_{end} \rangle$ where $s_{start}$ is the state of the agent at the beginning of motion and $s_{end}$ is the state of the agent at the end of  motion. For example, a state may be defined as a position and time $s=\langle x,y,t \rangle$ or it may also include velocity (and acceleration) components: $s=\langle x,y,\dot{x},\dot{y},t \rangle$; $s=\langle x,y,\dot{x},\dot{y},\ddot{x},\ddot{y},t \rangle$. The definition of states, including the number of dimensions will depend upon the application. A motion segment is continuous between $s_{start}$ and $s_{end}$, thus the transition between them must be kinematically feasible. Finally, a \emph{path} is composed of a sequence of states $\pi=[s_0,...,s_d]$, or alternately, a sequence of motions $\pi=[m_0,...,m_d]$ where each $s_{end}\in m_i = s_{start}\in m_{i+1}$.

\subsection{Geometric Containers}\label{static}

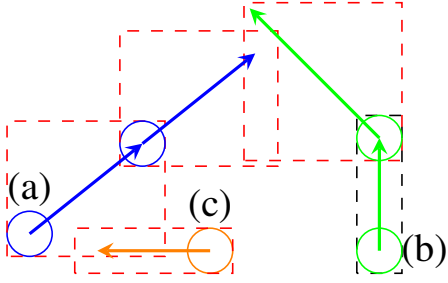
\begin{figure}
\centering
\scalebox{1.5}{
\begin{tikzpicture}

\draw [color=blue] (.9,.15) circle (.2);
\draw [dashed,color=red] (.7,-0.05) rectangle (2.1,1.15);
\draw [color=blue] (1.9,.95) circle (.2);
\draw [dashed,color=red] (1.7,.75) rectangle (3.1,1.95);
\draw [dashed] (3.8,-0.2) rectangle (4.2,1.2);
\draw [color=green] (4,0) circle (.2);
\draw [dashed,color=red] (2.8,.8) rectangle (4.2,2.2);
\draw [color=green] (4,1) circle (.2);
\draw [color=orange] (2.5,0) circle (.2);
\draw [dashed,color=red] (1.3,-.2) rectangle (2.7,.2);
\draw[->,>=stealth,thick,color=blue] (.9,.15) -- (1.9,.95);
\draw[->,>=stealth,thick,color=blue] (1.9,.95) -- (2.9,1.75);
\draw[->,>=stealth,thick,color=green] (4,0) -- (4,1);
\draw[->,>=stealth,thick,color=green] (4,1) -- (2.85,2.15);
\draw[->,>=stealth,thick,color=orange] (2.5,0) -- (1.5,0);

\node at (.9,.55) {(a)};
\node at (4.4,0) {(b)};
\node at (2.5,.4) {(c)};

\end{tikzpicture}
}
\caption{Collision detection using geometric containers. A collision is correctly detected between agents (a) and (b), but erroneously detected between (a) and (c).}
\label{fig:static}
\end{figure}

Geometric containers encapsulate portions of segmented motion in time and space using polygons, polytopes or spheres \cite{wagner2005geometric}. Then intersection detection is detected between the geometric containers of differing agents to determine if a collision has occurred. There are various approaches to intersection detection for
stationary objects \cite{jimenez20013d,kockara2007collision}.

In Figure \ref{fig:static} an example of this approach is shown which uses axis-aligned bounding boxes as geometric containers. The temporal dimensions are not  shown, but each bounding box also has a temporal component. Agent (a), (b) and (c) take actions (represented as directed edges) to arrive at their goal. Axis-aligned bounding boxes are reserved for each of these edges, then an intersection check is carried out. Although a collision is correctly detected between (a) and (b), an erroneous collision is detected between (b) and (c). Although this approach is computationally fast, it will reserve more temporospatial area than necessary, especially when long edges are present in a path, resulting in the possibility of type I errors.

\subsection{Incremental/Sampling-Based}
\label{incrementalsampling-based}

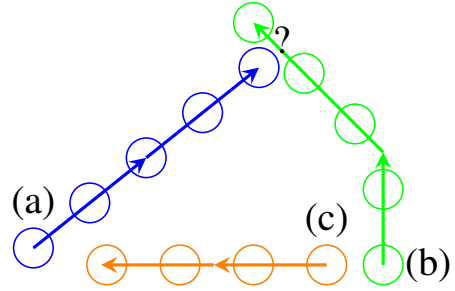
\begin{figure}
\centering
\scalebox{1.5}{
\begin{tikzpicture}

\draw [color=blue] (.9,.15) circle (5pt);
\draw [color=blue] (1.4,.55) circle (5pt);
\draw [color=blue] (1.9,.95) circle (5pt);
\draw [color=blue] (2.4,1.35) circle (5pt);
\draw [color=blue] (2.9,1.75) circle (5pt);

\draw [color=green] (4,0) circle (5pt);
\draw [color=green] (4,.67) circle (5pt);
\draw [color=green] (3.75,1.25) circle (5pt);
\draw [color=green] (3.3,1.7) circle (5pt);
\draw [color=green] (2.85,2.15) circle (5pt);

\draw [color=orange] (3.5,0) circle (5pt);
\draw [color=orange] (2.85,0) circle (5pt);
\draw [color=orange] (2.2,0) circle (5pt);
\draw [color=orange] (1.55,0) circle (5pt);

\draw[->,>=stealth,thick,color=blue] (.9,.15) -- (1.9,.95);
\draw[->,>=stealth,thick,color=blue] (1.9,.95) -- (2.9,1.75);
\draw[->,>=stealth,thick,color=green] (4,0) -- (4,1);
\draw[->,>=stealth,thick,color=green] (4,1) -- (2.85,2.15);
\draw[->,>=stealth,thick,color=orange] (3.5,0) -- (2.5,0);
\draw[->,>=stealth,thick,color=orange] (2.5,0) -- (1.5,0);

\node at (3.1,2) {?};
\node at (.9,.55) {(a)};
\node at (4.4,0) {(b)};
\node at (3.5,.4) {(c)};

\end{tikzpicture}
}
\caption{Sampling-based collision detection. A collision is not detected between agents (a) and (b).}
\label{fig:sampling}
\end{figure}

This approach involves translating objects along their trajectories incrementally and using static collision detection methods to detect overlaps at each increment. Figure \ref{fig:sampling} shows an example of this approach. Agents are translated to regular intervals along their trajectories, then intersection checks are performed at each interval. In contrast to the example in Figure \ref{fig:static}, there is no erroneous collision detected (type I error) between agent (a) and agent (c). However, a false negative (type II error) occurs between agent (a) and (b). The sampling approach is very important, samples too far apart may leave a  collision undetected, but samples very close together are computationally costly. Adaptive sampling approaches can help improve the accuracy and computational cost \cite{adaptive}.

In grid worlds, Brezenham's line algorithm \cite{brezenham}, a coarse form of collision detection can be used for selecting a specific set of grid-squares covered by a trajectory and then checking whether multiple agents are in the same grid square at intersecting times. A tighter approach based on Wu's antialiased line algorithm \cite{wu} is used in the AA-SIPP(m) \cite{AASIPP} algorithm. These methods may cause type I errors, but are guaranteed to avoid type II errors.

\subsection{Algebraic}\label{algebraic}

\begin{figure}
\centering
\includegraphics[width=\columnwidth,height=3.7cm,keepaspectratio=false]{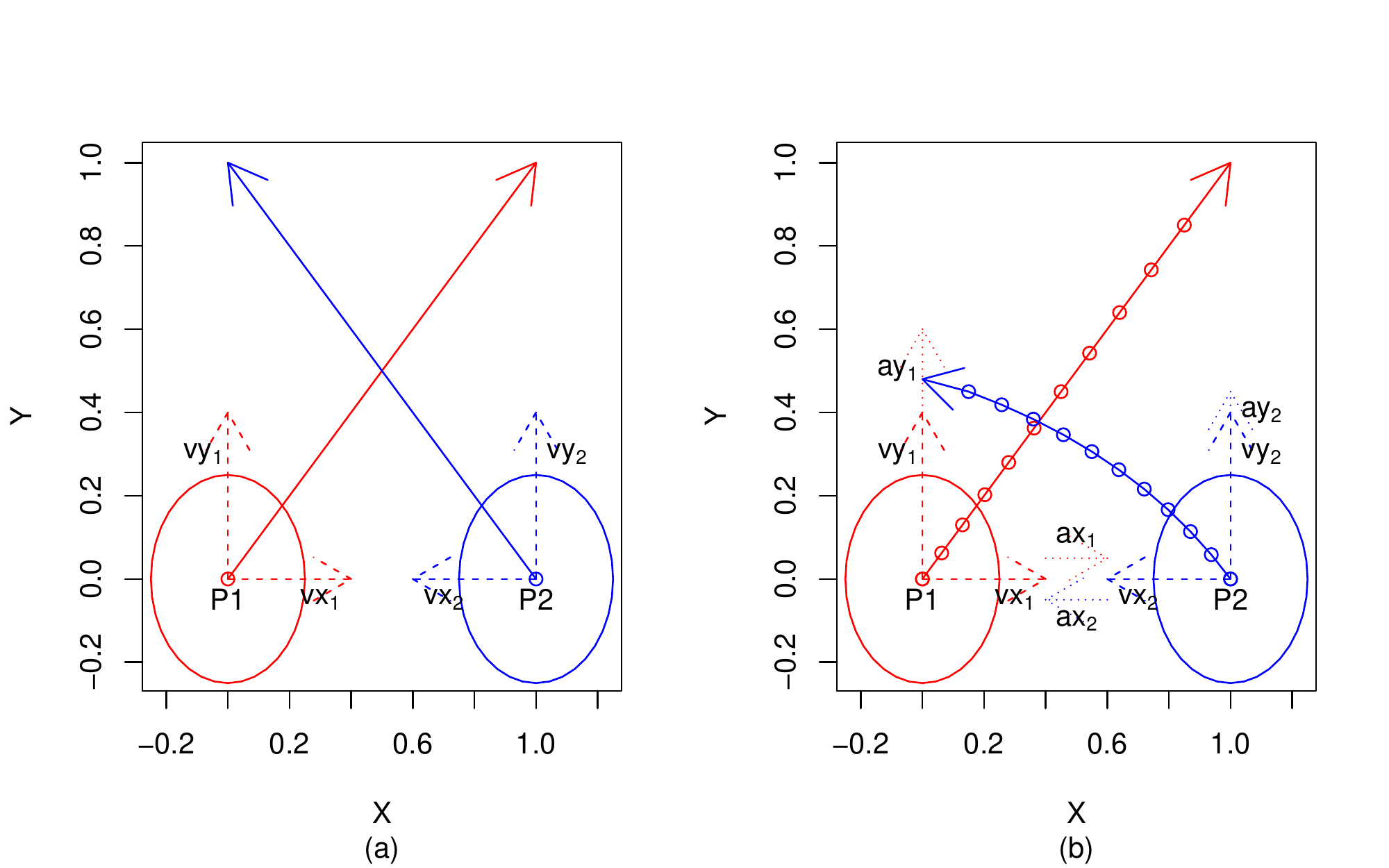}
\caption{\label{fig:both}Algebraic collision detection for trajectories with (a) constant velocity and (b) initial velocity with constant acceleration}
\end{figure}

By parameterizing the trajectory, closed-form solutions to
continuous-time conflict detection for circular, spherical, \cite{quadratic,ho2019multi} and triangular \cite{triangular} shaped agents have been formulated. An example for circular agents is shown in Figure \ref{fig:both}. In diagram (a), Agents move along their respective trajectories shown as a solid arrow. These trajectories are parameterized by the $x,y$ velocity vectors as shown with dashed arrows. Diagram (b) shows a similar scenario with acceleration vectors added to the tips of the velocity arrows shown with dotted arrows. Algebraic methods will calculate the exact time of collision between two moving agents assuming constant velocity and direction and also with acceleration.

When dealing with discrete-length motion segments, algebraic methods can be used to determine whether a collision will occur during the segment of motion. Deeper details of these calculations for circular agents are discussed in section \ref{closedform}.

\subsection{Geometric}\label{geometric}

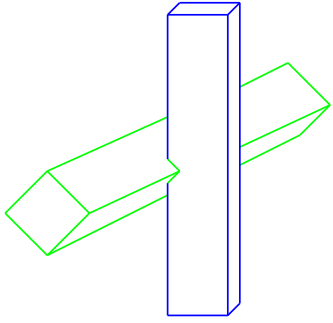
\begin{figure}
\label{fig:csg}
\centering
\scalebox{1.6}{
\begin{tikzpicture}

\draw[-,color=green] (1,1) -- (1.35,1.35);
\draw[-,color=green] (1,1) -- (.65,1.35);
\draw[-,color=green] (1,1.7) -- (1.35,1.35);
\draw[-,color=green] (1,1.7) -- (.65,1.35);

\draw[-,color=green] (1,1) -- (2,1.5);
\draw[-,color=green] (2,1.6) -- (2.1,1.7);
\draw[-,color=green] (1.35,1.35) -- (2.1,1.7);
\draw[-,color=green] (1,1.7) -- (2,2.15);
\draw[-,color=green] (2.1,1.7) -- (2,1.8);

\draw[-,color=green] (3.0,2.6) -- (2.6,2.4);
\draw[-,color=green] (3.0,2.6) -- (3.35,2.25);
\draw[-,color=green] (2.6,1.9) -- (3.35,2.25);
\draw[-,color=green] (3.1,2.) -- (3.35,2.25);
\draw[-,color=green] (3.1,2.) -- (2.6,1.75);

\draw[-,color=blue] (2,.5) -- (2.5,.5);
\draw[-,color=blue] (2,.5) -- (2,1.6);
\draw[-,color=blue] (2,1.8) -- (2,3);
\draw[-,color=blue] (2.5,.5) -- (2.5,3);
\draw[-,color=blue] (2,3) -- (2.5,3);
\draw[-,color=blue] (2.6,0.6) -- (2.6,3.1);
\draw[-,color=blue] (2.5,.5) -- (2.6,0.6);
\draw[-,color=blue] (2.5,3) -- (2.6,3.1);
\draw[-,color=blue] (2,3) -- (2.1,3.1);
\draw[-,color=blue] (2.1,3.1) -- (2.6,3.1);

\end{tikzpicture}
}
\caption{Constructive solid geometry collision detection. Time is \emph{extruded} into the model as an extra dimension, after which polygonal intersection detection is performed.}
\end{figure}

\input{VO.tex}

Geometric solutions are the most computationally expensive collision
detection approaches, however they are formulated for many different
obstacle shapes - typically primitive shapes, polygons or meshes. Two of
the most popular approaches are constructive solid geometry (CSG)
\cite{CSG}, and velocity obstacles (VO) \cite{VO}.

CSG approaches treat the time domain as an additional polygonal
dimension, \emph{extruding} polygons into the time dimension, after
which a static polygonal intersection check is applied. Computation of
the extruded volumes can be very expensive and formulating ways to
enhance CSG has been a subject of ongoing research \cite{dyllong2007verified,kiel2013verified}.

Velocity obstacles have been formulated for infinite length vector collision detection for arbitrary-shaped agents \cite{VO}. A velocity obstacle is depicted in Figure \ref{fig:VO}. A VO is created for two agents \(A\) and \(B\), located with center points $A$ and $B$ as shown in diagram (a). The agents have shapes -- here shown as circles with radius $r_A$ and $r_B$. The agents' motion follows velocity vectors \(VA\) and \(VB\) shown as arrows.

In order to construct the VO, first, the shape of agent \(B\) is inflated by computing the Minkowski sum \(A\oplus B\) of the two agent's shapes. Next, two tangent lines from point \(A\) to the sides of \(A\oplus B\) are calculated to form a polygon. Finally, the polygon is translated so that its apex is at $A+VB$. The area between the  translated tangent lines is the velocity obstacle (labeled VO in the diagram). The VO represents the unsafe region of velocity for agent $A$, assuming agent $B$ does not change it's trajectory. If the point $A+VA$ lies inside the VO, agent \(A\) will collide with agent \(B\) some time in the infinite future.

In the case of segmented motion, VOs can still be used for collision detection with some adaptations \cite{andreychuk2019continuous}. In addition, collision avoidance can be achieved by choosing a velocity for $A$ such that $A+VA$ lies outside the VO. One approach is to set $VA$ so that $A+VA$ lies on the intersection point of either of the VO tangent lines $\pm \epsilon$.

\subsection{Summary}\label{summary}

Depending on the application, any of the above methods may meet the problem constraints.
Static detection is the approach of choice for domains with
discretized-time movement models as it is the cheapest and (depending on
the movement model) may yield no loss in accuracy. In continuous-time
domains, one of the latter choices is usually preferable, with sampling
often being the cheapest approach, followed by algebraic and geometric
approaches. There is a trade-off with respect to accuracy and
computational cost. The latter approaches provide the most flexibility
when high accuracy and complex agent shapes are necessary.

\section{Closed-Form Collision Detection for Circular
Agents}\label{closedform}

Figure \ref{fig:both} shows an example of two-agent motion for fixed
velocity (a) and initial velocity with fixed acceleration (b). Computing
the time and duration of conflict for two circular agents can be done by
solving equations for the squared distance between agents (Ericson
(2004)).

\subsection{Constant Velocity}\label{constant-velocity}

Given \(P_1=\langle x_1,y_1 \rangle\), the start position of agent \(1\), and
\(P_2=\langle x_2,y_2 \rangle\), the start position of agent \(2\), velocity vectors
\(V_1=\langle vx_1,vy_1 \rangle\), \(V_2=\langle vx_2,vy_2 \rangle\), and radii \(r_1\), \(r_2\)
respectively, the location in time of an agent is defined as:

\begin{equation} \label{eq:translate}
P'=P+Vt
\end{equation}

The following equation specifies the squared distance between the centers of the agents over time:

\begin{equation} \label{eq:ctime1}
sqdist(t)={V_\Delta}^2t^2+2V_\Delta\boldsymbol{\cdot} P_\Delta t+{P_\Delta}^2
\end{equation}

where

$ $

\(P_\Delta=P_1-P_2\)

\(V_\Delta=V_1-V_2\)

$ $

Via substitution, this equation is simplified to a quadratic equation:

\begin{equation} \label{eq:ctime2}
sqdist(t)=at^2+bt+c_0
\end{equation}

where

$ $

\(a={V_\Delta}^2\)

\(b=2V_\Delta\boldsymbol{\cdot} P_\Delta\)

\(c_0={P_\Delta}^2\)

$ $

A collision will occur when the squared distance between the agents is less than or equal to the squared sum of the radii, giving the following inequality. 

$$at^2+bt+c_0\leq (r_1+r_2)^2$$

Solving the inequality gives the equation for collision between the agent's edges:

$$0\geq at^2+bt+c_0-(r_1+r_2)^2$$

\begin{equation} \label{eq:ctime3}
sqEdgeDist(t)=at^2+bt+c
\end{equation}

where

$ $

\(c=P_{\Delta}^2-(r_1+r_2)^2\)

$ $

Solving equation \ref{eq:ctime3} for $t$ will determine the exact times where the squared distance between agent's edges is zero -- the time when collision occurs. Section \ref{computing-the-exact-conflict-interval} discusses the process for determining the conflict interval for using this equation.

\subsection{Initial Velocity with Constant
Acceleration}\label{initial-velocity-with-constant-acceleration}

Equation \(\eqref{eq:ctime3}\) can be extended for constant
acceleration. Given \(P_1=\langle x_1,y_1 \rangle\), the start position of agent
\(1\), and \(P_2=\langle x_2,y_2 \rangle\), the start position of agent \(2\),
velocity vectors \(V_1=\langle vx_1,vy_1 \rangle\), \(V_2=\langle vx_2,vy_2 \rangle\), acceleration
vectors \(A_1=\langle ax_1,ay_1 \rangle\), \(A_2=\langle ax_2,ay_2 \rangle\) and radii \(r_1\),
\(r_2\) respectively, the location in time of an agent is defined as:

\begin{equation} \label{eq:translate2}
P'=P+Vt+\frac{At^2}{2}
\end{equation}

The following inequality specifies the collision condition as a quartic equation:

\begin{equation} \label{eq:accel}
at^4 + bt^3 + ct^2 + dt + e_0 \leq (r_1+r_2)^2
\end{equation}

$ $

where

$ $

\(a=\frac{{A_\Delta}^2}{4}\)

\(b={A_\Delta\boldsymbol{\cdot}V_\Delta}\)

\(c={A_\Delta\boldsymbol{\cdot} P_\Delta + V_\Delta}^2\)

\(d=2V_\Delta\boldsymbol{\cdot} P_\Delta\)

\(e_0={P_\Delta}^2\)

$ $

for

$ $

\(P_\Delta=P_1-P_2\)

\(V_\Delta=V_1-V_2\)

\(A_\Delta=A_1-A_2\)

$ $

which gives the equation for the squared distance between circular
edges:

\begin{equation} \label{eq:accel2}
sqEdgeDist(t)=at^4 + bt^3 + ct^2 + dt + e
\end{equation}

where

$ $

\(e={P_\Delta}^2-(r_1+r_2)^2\)

$ $

Again, solving for $t$ will yield the time of collision, which is discussed further in the next section.

\section{Computing the Exact Conflict
Interval}\label{computing-the-exact-conflict-interval}

The exact conflict interval is determined by solving for the roots of
\(\eqref{eq:ctime3}\) or \(\eqref{eq:accel2}\) using the quadratic and
quartic formulas respectively. These solutions assume that both agents
are at \(P_1\) and \(P_2\) at the same time. However, if there is an
offset in time, e.g.~agent 1 starts moving at time \(t_1\) and agent 2
starts moving at time \(t_2\), then \(P_\Delta\) must be adjusted to
reflect this offset by projecting the position of the earlier agent to
be at the position when the later agent starts its motion. If the earlier agent were agent 1, the adjustment would be as follows:

\begin{equation} \label{eq:adjust}
P_\Delta=P_1+V_1(t_2-t_1)-P_2
\end{equation}

Otherwise, the adjustment will be analogously done for agent 2. In the
case of acceleration, the position and velocity must be adjusted (again,
assuming agent 1 starts early) as:

\begin{equation} \label{eq:adjust2}
P_\Delta=P_1+V_1(t_2-t_1)+\frac{A_1(t_2-t_1)^2}{2}-P_2
\end{equation}

\begin{equation} \label{eq:adjust3}
V_\Delta=V_1+A_1(t_2-t_1)-V_2
\end{equation}

\subsection{Constant Velocity}\label{constant-velocity-1}

\begin{figure}
\centering
\includegraphics[width=\columnwidth,height=4.2cm,keepaspectratio=false]{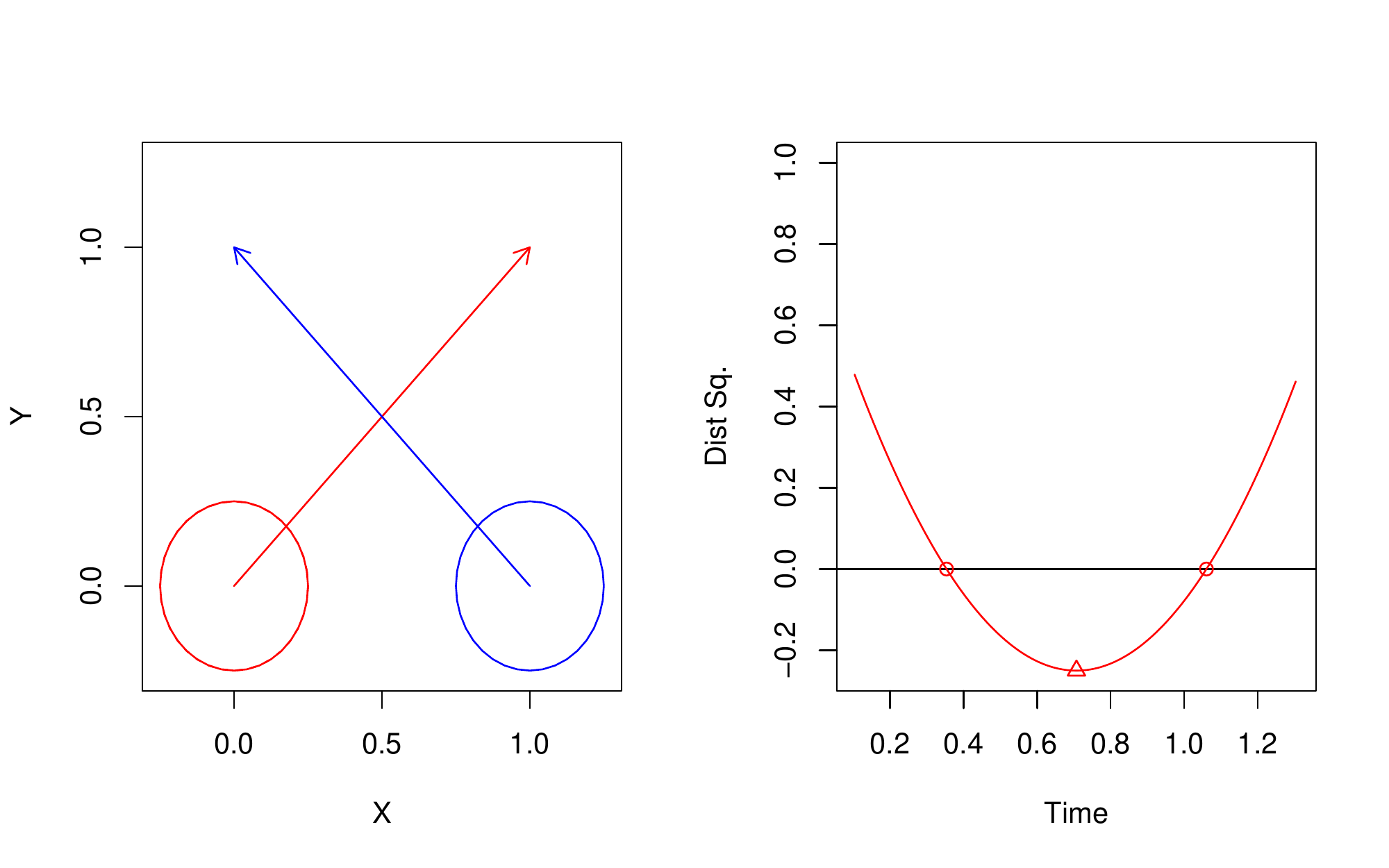}
\caption{\label{fig:quad}Agents Trajectories and Corresponding Squared
Distance Plot}
\end{figure}

For the quadratic form, if the discriminant (\(b^2-4ac\)) is less than
zero, \(V_1\) and \(V_2\) are parallel and no collision will ever occur.
Assuming the discriminant is positive, the collision interval is defined
as the roots of the quadratic formula:

\begin{equation} \label{eq:quadratic}
t_{interval}=\frac{-b\pm \sqrt{b^2-4ac}}{2a}
\end{equation}

In the case of a double root, the edges of the agents just touch, but no
overlap actually occurs (assuming open intervals). See Figure \ref{fig:quad} for an example of two-agent motion and the resulting squared-distance plot. When the distance is less than zero, there is overlap of the agents. Given this interval, it is possible to determine whether a collision will occur in the future and at what time, or if the agents are currently colliding.

\subsection{Initial Velocity with Constant
Acceleration}\label{initial-velocity-with-constant-acceleration-1}

This case uses the quartic formula to find roots to
\(\eqref{eq:accel2}\). The quartic formula will yield 4 roots, some of
which may be imaginary resulting in 0, 1 or 2 conflict intervals.
Imaginary roots will tell us the time(s) at which agents are locally
closest together, but do not actually overlap (local minima). Imaginary
roots are always double roots, and can be discarded. If all 4 roots are
imaginary, the agents never overlap. If there is a double real root,
then the two agents touch edges at exactly one point in time, creating
an instantaneous interval.

Because our equation is based on distance,
the quartic function will always be concave up. Hence, the overlapping
intervals can only be between roots 1,2 and 3,4. If roots 1,2 and/or 3,4
are real, then the agents continuously overlap between 1,2 and/or 3,4
respectively. Four real roots means that the objects overlap twice,
continuously between root pairs 1,2 and 3,4. This is possible because
agents may have curved trajectories. See Figure \(\ref{fig:both}\) (b)
for an example.

\section{Determining Exact Minimum Delay or Velocity Adjustment for Conflict Avoidance}\label{determining-exact-minimum-delay-for-conflict-avoidance}

It is often useful, not just to determine if and when agents are going
to collide, but to determine a delay time to avoid collision.

\subsection{Exact Delay for Constant Velocity}\label{constant-velocity-2}

In order to determine the minimum delay required for an agent to avoid
conflict, we adjust (3) to incorporate \(\delta=t_2-t_1\), a delay variable, by
plugging equation \(\eqref{eq:adjust}\) into equation \(\eqref{eq:ctime1}\) to
get:

\begin{equation} \label{eq:ellipse}
sqEdgeDist(t,\delta)=At^2+Bt\delta+C\delta^2+Dt+E\delta+F
\end{equation}

where

$ $

\(A={V_\Delta}^2\)

\(B=2(V_1^2 - V_1\boldsymbol{\cdot}V_2)\)

\(C={V_1}^2\)

\(D=2(P_2\boldsymbol{\cdot}V_2 - P_2\boldsymbol{\cdot}V_1 + P_1\boldsymbol{\cdot}V_2 - P_1\boldsymbol{\cdot}V_1)\)

\(E=-2(P_2\boldsymbol{\cdot}V_1+P_1\boldsymbol{\cdot}V_1)\)

\(F={V_\Delta}^2 - (r_1+r_2)^2\)

$ $

Equation \(\eqref{eq:ellipse}\) is the standard form of a conic section. Note that the sign of both \(A\) and \(C\) are positive, therefore, this conic section will always be an ellipse, except for two degenerate cases: (1) agents' motion is parallel and (2) at least one agent is waiting in place. Fortunately, both cases are easy to detect and solve. The conversion of (8) to canonical form for an ellipse will not be covered here, nor is it
necessary.

\begin{figure}

\centering
\includegraphics[width=\columnwidth]{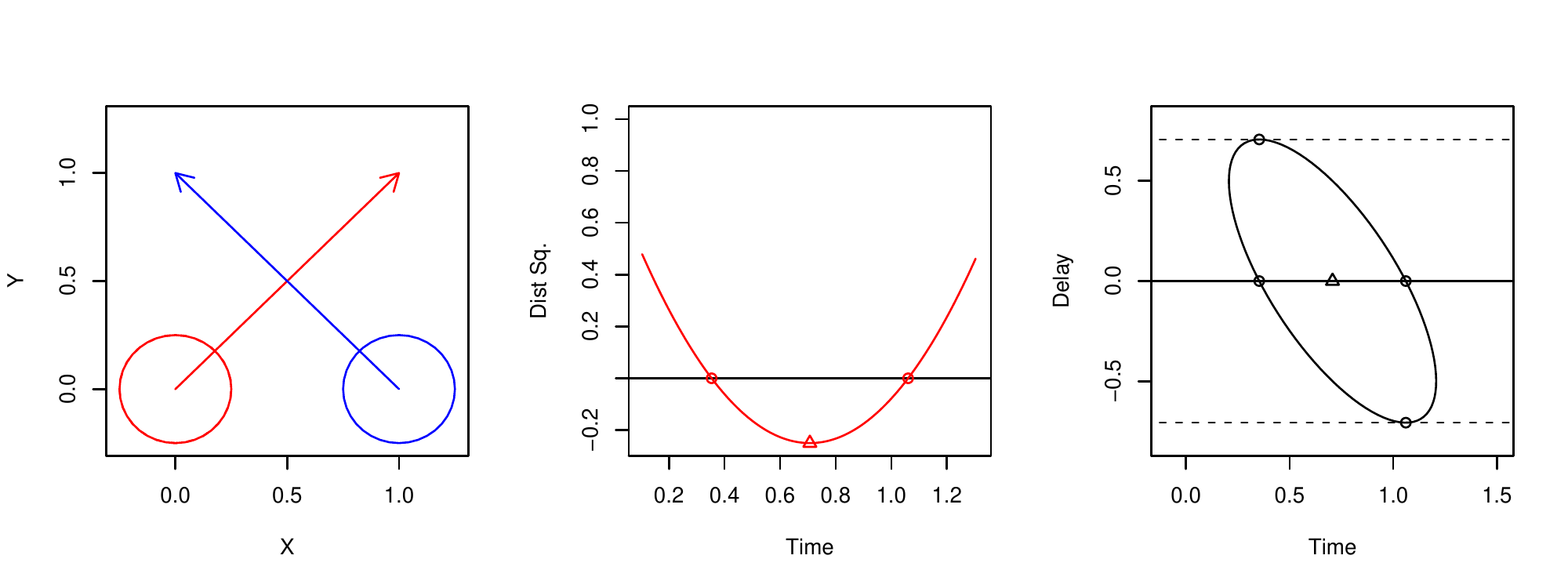}

(a)

\includegraphics[width=\columnwidth]{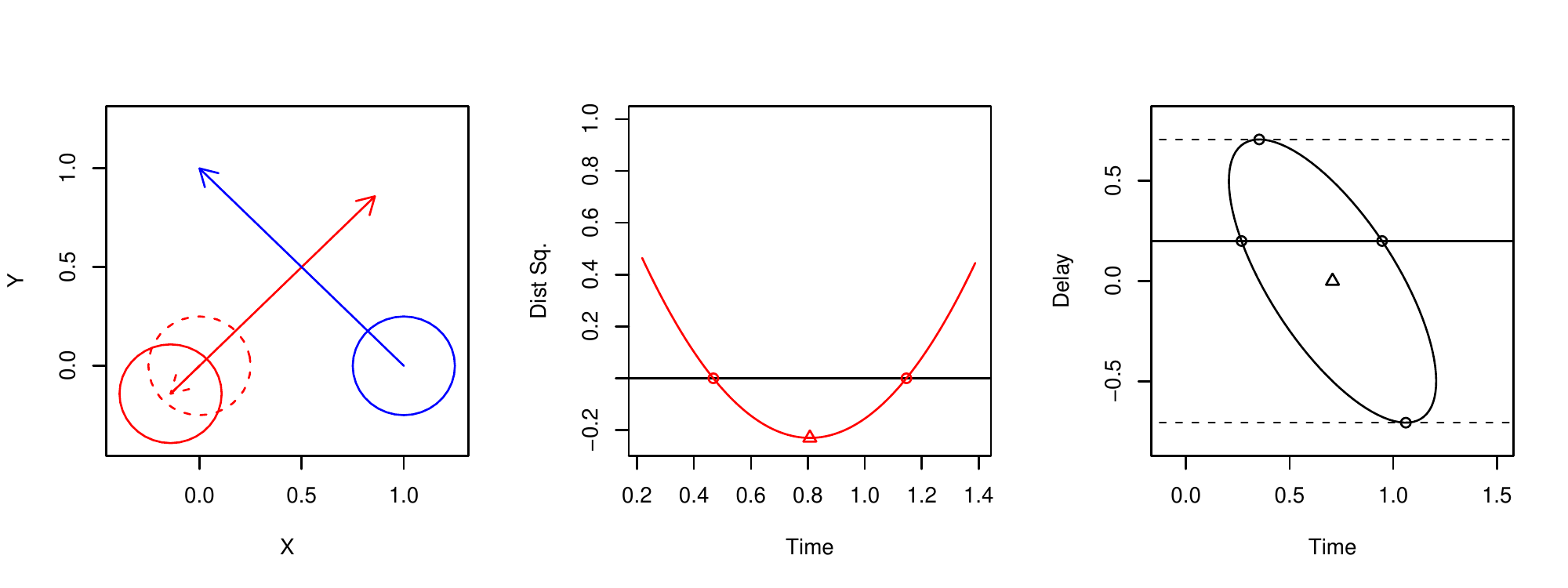}
(b)
\caption{\label{fig:conic} (a) Agent trajectories, squared distance plot and
ellipse showing collision intervals for Varying \(delay\)
and (b) the same trajectories where the red agent is delayed delayed by 0.2 seconds}
\end{figure}

Figure \ref{fig:conic}(a) shows an example of agent trajectories, the squared distance plot (equation \eqref{eq:ctime3}) when \(delay=0\), and the resulting conic section (equation \eqref{eq:ellipse}). Note that the horizontal line at \(\delta=0\) passes through the ellipse at the exact same time points that the squared distance plot does. If agent 1 were to delay by \(\epsilon\), the horizontal line would move up, resulting in a different collision interval (see Figure \ref{fig:conic}(b)). If agent 2 were to delay by \(\epsilon\), the horizontal line would move down, again resulting in a different collision interval. The question we want to solve is: what value of \(delay\) will result in no collision? In other words, we want to find the positive value of \(\delta\), such that the radii of the
agents just touch, i.e. \eqref{eq:ellipse} yields a double root. The
targeted delay interval is derived by determining the top and bottom
extrema of the ellipse \cite{hendricks2012rotated}.

\begin{equation} \label{eq:delayrange}
\resizebox{.9 \columnwidth}{!} 
{%
$delayRange=center_\delta\pm\frac{\sqrt{(2BD-4AE)^2 + 4(4AC-B^2)(D^2-4AF)}}{2(4AC-B^2)}$%
}
\end{equation}

where \(center_\delta\) is the y-coordinate of the ellipse center:

$ $

\(center_\delta=\frac{BD-2AE}{4AC-B^2}\)

$ $

The collision times of the endpoints of the delayRange are computed via:

\begin{equation} \label{eq:extrema}
collisionTimes=\frac{-B(delayRange)-D}{2A}
\end{equation}

Note that \eqref{eq:delayrange} is undefined when the discriminant is
negative, which can only happen for $a=0$ or $c=0$. This can only happen
when agents' motion vectors are parallel (moving the same or opposite
directions) or either agent is waiting in place. These cases are easy to
detect.

\begin{figure}[b]
\centering
\includegraphics[width=\columnwidth,height=3cm]{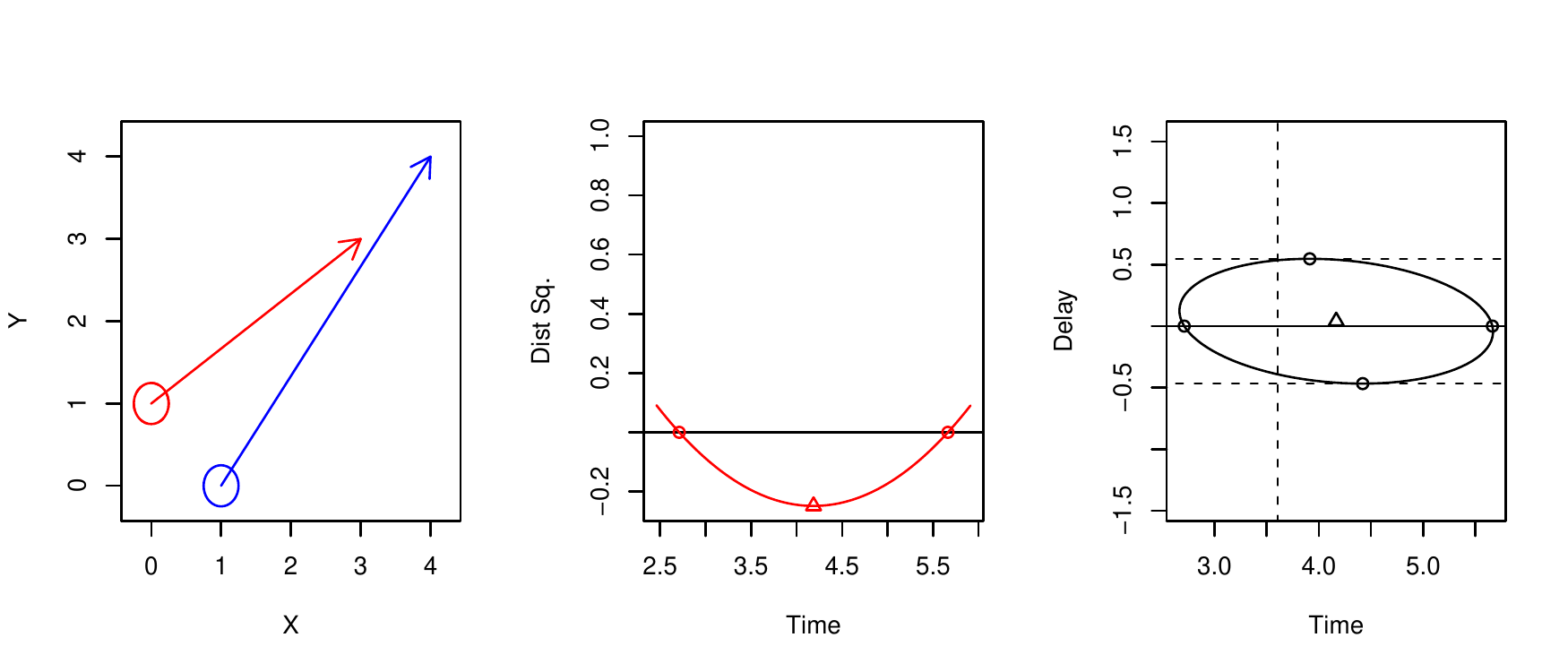}
\caption{\label{fig:maxdelay}An example where the maximum delay time
happens after the first agent arrives at its destination.}
\end{figure}
When the motion is not of infinite length, i. e. segmented motion, we must also
take into account the beginning and end of the duration of motion.
Effectively, we treat agents as if they appear at their start time and
disappear at their end time. When the movement of agents 1 and 2 start
at \(t_1\) and \(t_2\) and end at \(t_1'\) and \(t_2'\) respectively, we
measure time relative to \(t_0=MIN(t_1,t_2)\) and
\(t_{max}=MIN(t_1',t_2')\). In the case that \(\delta=t_1-t_2\) is outside of
the range \(delayRange\) as calculated via \eqref{eq:delayrange}, no
collision will occur. If either of the collision times (as calculated in
\eqref{eq:extrema} for each point in \(delayRange\) occur before
\(t_0\), or after \(t_{max}\), the delay times need to be re-computed
for \(t_0\) or \(t_{max}\) as necessary using \eqref{eq:delta}. An
example where \(t_{max}\) occurs too early is shown by the vertical
dashed line in Figure \ref{fig:maxdelay}.

This yields the algorithm detailed in Algorithm \ref{alg:interval} for computing the unsafe interval for segmented motion. The algorithm is straightforward and utilizes the following additional formulas:

The value of \(\delta\), given a time which is derived from
\eqref{eq:ellipse}, solved for \(\delta\):

\begin{equation} \label{eq:delta}
\delta = \frac{-\sqrt{(Bt + E)^2 - 4C (t(At + D) + F)} + Bt + E}{2C}
\end{equation}

The leftmost \(t\) coordinate on the ellipse:

\begin{equation} \label{eq:leftmost}
\resizebox{.9 \columnwidth}{!} 
{%
$minCollisionTime=center_t-\frac{\sqrt{(2BE-4CD)^2 + 4(4AC-B^2)(E^2-4CF)}}{2(4AC-B^2)}$%
}
\end{equation}

where \(center_t\) is \(\frac{BE-2CD}{4AC-B^2}\)

At Algorithm \ref{alg:interval}, lines 5-9 check the actual delay ($\delta=t2-t1$) between the two agents against the unsafe delay range per equation \eqref{eq:delayrange}. If there is no collision (e.g. in the case of parallel movement) or $\delta$ does not fall inside the unsafe range, no collision will occur. Lines 10-23 compute the unsafe time interval per equation \eqref{eq:extrema} and then adjust the endpoints accordingly per t0 and tmax using equation \eqref{eq:delta}.

The final result is the adjusted unsafe interval for agent 1. This interval can now be used to instruct agent 1 not to start execution of its action inside the interval (e.g. by starting its action sooner or later). Note that the unsafe interval for agent 2 is the negated interval for agent 1 -- [-range[2],-range[1]].

\begin{algorithm}[t]
\caption{Unsafe Interval Computation for Segmented Motion}\label{alg:interval}
\begin{algorithmic}[1]
  \State INPUT: P1,P2,V1,V2,t1,t2,t1',t2',r1,r2
  \State t0=MAX(t1,t2)
  \State tmax$\gets$MIN(t1',t2')
  \State $\delta$=t2-t1
  \State // Execute equation \eqref{eq:delayrange} to get unsafe delay range
  \State range$\gets$delayRange(P1,P2,V1,V2,r1,r2)
  \If{range=$\emptyset$ or range[1]$>\delta$ or range[2]$<\delta$}
    \State \textbf{return} NO COLLISION
  \EndIf
  \State // Execute equation \eqref{eq:extrema} to get unsafe time range
  \State collisionTimes$\gets$delayTimes(P1,P2,V1,V2,r1,r2)
  \State minCollisionTime$\gets$MIN(collisionTimes)
  \State maxCollisionTime$\gets$MAX(collisionTimes)

  \State // Truncate delay for motion time segments
  \If{minCollisionTime$<$t0}
    \State // Get delay for t0 via \eqref{eq:delta}
    \State range[1]$\gets$delayAtTime(P1,P2,V1,V2,r1,r2,t0)
  \EndIf
  \If{maxCollisionTime$<$t0}
    \State // Get delay for tmax via \eqref{eq:delta}
    \State range[2]$\gets$delayAtTime(P1,P2,V1,V2,r1,r2,t0)
  \EndIf
  
  \State // Return the unsafe interval by adding the delay to the start time
  \State \textbf{return} [t0+range[1],MIN(tmax,t0+range[2])]
\end{algorithmic}
\end{algorithm}

\subsection{Exact Delay for Initial Velocity with Constant
Acceleration}\label{initial-velocity-with-constant-acceleration-2}

The equivalent conic equation for 4th order bivariates is called a
quartic plane curve. A closed-form solution for unsafe intervals is still an open question. However, an interative solution has been formulated for
the constant velocity case which is generalizable to this case
\cite{andreychuk2019continuous}.

The algorithm starts by evaluating \eqref{eq:accel2} at \(t0\),
retrieving an initial upper bound from the interval which is closest to
and greater than \(t0\). Then performs a binary search, from both ends of
the interval until the interval is determined within a predetermined accuracy threshold. Binary search is a
well known algorithm and will not be repeated here.

\subsection{Minimum Velocity Change for Constant Velocity}

\input{VO1.tex}

In order to determine the minimum velocity change necessary to avoid collision for segmented motion, a VO is created as shown in Figure \ref{fig:VO1} which is similar to Figure \ref{fig:VO}, but with motion segments added. Motion segments are shown as dotted arrows with large points at the beginning and end of the segment. Velocities that lie on the segment are the only valid choices, hence a velocity that lies just outside of the VO as shown in diagram (b) is desirable for determining the minimum necessary change to avoid collision. There may be kinematic constraints on agents, such as a maximum velocity.

The following steps can be undertaken to determine the appropriate action for the agent, which may result in the agent waiting in place or using a new velocity:

\begin{enumerate}
\item Detect if a collision will occur inside the segments. This can be done via equation \eqref{eq:ctime3}.

\begin{itemize}
\item Return if no collision
\end{itemize}
\item Construct a VO, then compute a new velocity that lies on the segment and intersects with the edges of the VO as shown in Figure \ref{fig:VO1} (b) for agent A. Ths can be done using a formula for the line intersection point \cite{antonio1992faster} of the motion vector and both of the VO tangent lines.
\begin{itemize}
\item Return new velocity if either of the velocities at the intersection points are kinematically feasible.
\end{itemize} 
\item Construct and check a VO for a new velocity for agent B.
\begin{itemize}
\item Return new velocity if either of the velocities at the intersection points are kinematically feasible.
\end{itemize}

\item If the current state of the agent will allow it to wait in place, compute the delay using Algorithm \ref{alg:interval}.
\begin{itemize}
\item Return original velocity and new delay.
\end{itemize}
\item Otherwise, return NO SOLUTION
\end{enumerate}

\section{Conclusion}\label{conclusion}

In this paper, we have provided an overview of collision detection for
polygonal and circular agents. We have also provided derivations for
computing the exact interval of collision between two agents with
constant velocity or intial velocity with constant acceleration. We have
additionally derived a formulation for computing unsafe intervals
(the range of start times in which agents come into collision) for two
circular agents with constant velocity and differing start times. An
algorithm was then shown for computing the unsafe intervals in the case
of segmented motion. Finally, an algorithm for computing safe velocities and delay times was outlined.

Future work may involve derivations of the exact formulation of unsafe
intervals for agents with acceleration.

\newpage

\bibliographystyle{aaai}
\bibliography{allrefs}

\end{document}

%% file: VO.tex
\begin{figure}
\label{fig:VO}
\centering
\scalebox{1.2}{
\begin{tikzpicture}

\draw [color=green,fill=green!30] (0.5,2.5) circle (10pt);
\draw [color=blue,fill=blue!30] (0.5,0.5) circle (10pt);
\draw[->,>=stealth,thick,color=blue] (.5,0.5) -- (2,1.0);
\draw[->,>=stealth,thick,color=green] (0.5,2.5) -- (2,2);

\node at (1.5,-.3) {(a)};
\node at (1.1,.5) {\tiny $\overrightarrow{VA}$};
\node at (1.1,2.5) {\tiny $\overrightarrow{VB}$};
\node at (.35,0.5) {\small $A$};
\node at (0.35,2.5) {\small $B$};

\end{tikzpicture}
}\scalebox{1.2}{
\begin{tikzpicture}
\draw [color=green] (2,0) circle (2pt);
\draw[fill=blue!25] (2,0) -- (1,2.5) -- (3,2.5) -- cycle;
\draw[color=white,fill=gray!25] (.5,.5) -- (-.5,3) -- (1.5,3) -- cycle;
\draw [color=black,dashed,] (0.5,2.5) circle (21pt);
\draw [color=gray,dashed,] (2,2) circle (21pt);
\draw [color=green,fill=green!30] (0.5,2.5) circle (10pt);
\draw [color=blue,fill=blue] (0.5,0.5) circle (2pt);
\draw [color=blue] (2,1) circle (2pt);

\draw[->,>=stealth,dashed,color=black] (0.5,.5) -- (-.5,3);
\draw[->,>=stealth,dashed,color=black] (0.5,.5) -- (1.5,3);
\draw[->,>=stealth,thick,color=blue] (.5,0.5) -- (2,1.0);
\draw[->,>=stealth,thick,color=green] (0.5,.5) -- (2,0);
\node at (1.5,-.3) {(b)};
\node at (2.5,1) {\tiny $A\!+\!\overrightarrow{VA}$};
\node at (2.5,0) {\tiny $A\!+\!\overrightarrow{VB}$};
\node at (.2,0.5) {\small $A$};
\node at (2,1.6) {\small $VO$};
\node at (0.75,2.5) {\small $B$};
\node[rotate=45] at (0.2,2.45) {\tiny $r_A\!+\!r_B$};
\draw[<->,>=stealth,very thin,color=black] (0.55,2.45) -- (0,2);

\end{tikzpicture}
}
\caption{ Velocity Obstacle (VO) construction based on (a) two agents with motion vectors. The trajectories and shapes of agents are interpreted to create (b) the velocity obstacle -- labeled 'VO`}
\end{figure}

%% file: VO1.tex
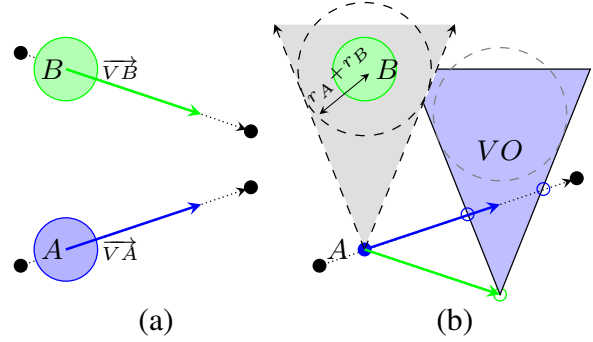
\begin{figure}
\centering
\scalebox{1.2}{
\begin{tikzpicture}
\draw [color=black,fill=black] (0,2.68) circle (2pt);
\draw [color=black,fill=black] (2.55,1.81) circle (2pt);
\draw[->,>=stealth,densely dotted,color=black] (0,2.68) -- (2.5,1.83);
\draw [color=black,fill=black] (0,.32) circle (2pt);
\draw [color=black,fill=black] (2.55,1.19) circle (2pt);
\draw[->,>=stealth,densely dotted,color=black] (0,.32) -- (2.5,1.17);
\draw [color=green,fill=green!30] (0.5,2.5) circle (10pt);
\draw [color=blue,fill=blue!30] (0.5,0.5) circle (10pt);
\draw[->,>=stealth,thick,color=blue] (.5,0.5) -- (2,1.0);
\draw[->,>=stealth,thick,color=green] (0.5,2.5) -- (2,2);

\node at (1.5,-.3) {(a)};
\node at (1.1,.5) {\tiny $\overrightarrow{VA}$};
\node at (1.1,2.5) {\tiny $\overrightarrow{VB}$};
\node at (.35,0.5) {\small $A$};
\node at (0.35,2.5) {\small $B$};

\end{tikzpicture}
}\scalebox{1.2}{
\begin{tikzpicture}
\draw [color=green] (2,0) circle (2pt);
\draw[fill=blue!25] (2,0) -- (1,2.5) -- (3,2.5) -- cycle;
\draw[color=white,fill=gray!25] (.5,.5) -- (-.5,3) -- (1.5,3) -- cycle;
\draw[->,>=stealth,densely dotted,color=black] (0,.32) -- (2.8,1.27);
\draw [color=black,dashed,] (0.5,2.5) circle (21pt);
\draw [color=gray,dashed,] (2,2) circle (21pt);
\draw [color=green,fill=green!30] (0.5,2.5) circle (10pt);
\draw [color=blue,fill=blue] (0.5,0.5) circle (2pt);

\draw[->,>=stealth,dashed,color=black] (0.5,.5) -- (-.5,3);
\draw[->,>=stealth,dashed,color=black] (0.5,.5) -- (1.5,3);
\draw[->,>=stealth,thick,color=blue] (.5,0.5) -- (2,1.0);
\draw[->,>=stealth,thick,color=green] (0.5,.5) -- (2,0);
\node at (1.5,-.3) {(b)};
\node at (.2,0.5) {\small $A$};
\node at (2,1.6) {\small $VO$};
\node at (0.75,2.5) {\small $B$};
\node[rotate=45] at (0.2,2.45) {\tiny $r_A\!+\!r_B$};
\draw[<->,>=stealth,very thin,color=black] (0.55,2.45) -- (0,2);
\draw [color=blue] (1.64,.89) circle (2pt);
\draw [color=blue] (2.48,1.17) circle (2pt);
\draw [color=black,fill=black] (0,.32) circle (2pt);
\draw [color=black,fill=black] (2.85,1.29) circle (2pt);
\end{tikzpicture}
}
\caption{\label{fig:VO1} Velocity Obstacle (VO) construction based on (a) two agents moving on edges. (b) The minimum change for safe velocity is determined by the intersection points of the edge and the velocity obstacle.}
\end{figure}

%% file: MutuallyConflictingActionTimes.bbl
\begin{thebibliography}{}

\bibitem[\protect\citeauthoryear{Andreychuk \bgroup et al\mbox.\egroup
  }{2019}]{andreychuk2019continuous}
Andreychuk, A.; Yakovlev, K.; Atzmon, D.; and Stern, R.
\newblock 2019.
\newblock Multi-agent pathfinding with continuous time.
\newblock In {\em Proceedings of the Twenty-Eighth International Joint
  Conference on Artificial Intelligence, {IJCAI-19}}.
\newblock International Joint Conferences on Artificial Intelligence
  Organization.

\bibitem[\protect\citeauthoryear{Antonio}{1992}]{antonio1992faster}
Antonio, F.
\newblock 1992.
\newblock Faster line segment intersection.
\newblock In {\em Graphics Gems III (IBM Version)}. Elsevier.
\newblock  199--202.

\bibitem[\protect\citeauthoryear{Bresenham}{1987}]{brezenham}
Bresenham, J.~E.
\newblock 1987.
\newblock Ambiguities in incremental line rastering.
\newblock {\em IEEE Computer Graphics and Applications} 7(5):31--43.

\bibitem[\protect\citeauthoryear{Dyllong and Grimm}{}]{dyllong2007verified}
Dyllong, E., and Grimm, C.
\newblock Verified adaptive octree representations of constructive solid
  geometry objects.
\newblock Citeseer.

\bibitem[\protect\citeauthoryear{Ericson}{2004}]{quadratic}
Ericson, C.
\newblock 2004.
\newblock {\em Real-time collision detection}.
\newblock CRC Press.

\bibitem[\protect\citeauthoryear{Fiorini and Shiller}{1998}]{VO}
Fiorini, P., and Shiller, Z.
\newblock 1998.
\newblock Motion planning in dynamic environments using velocity obstacles.
\newblock {\em The International Journal of Robotics Research} 17(7):760--772.

\bibitem[\protect\citeauthoryear{Gilbert and Hong}{1989}]{adaptive}
Gilbert, E.~G., and Hong, S.
\newblock 1989.
\newblock A new algorithm for detecting the collision of moving objects.
\newblock In {\em Robotics and Automation, 1989. Proceedings., 1989 IEEE
  International Conference on},  8--14.
\newblock IEEE.

\bibitem[\protect\citeauthoryear{Hendricks}{2012}]{hendricks2012rotated}
Hendricks, M.~C.
\newblock 2012.
\newblock Rotated ellipses and their intersections with lines.

\bibitem[\protect\citeauthoryear{Ho \bgroup et al\mbox.\egroup
  }{2019}]{ho2019multi}
Ho, F.; Salta, A.; Geraldes, R.; Goncalves, A.; Cavazza, M.; and Prendinger, H.
\newblock 2019.
\newblock Multi-agent path finding for uav traffic management.
\newblock In {\em Proceedings of the 18th International Conference on
  Autonomous Agents and MultiAgent Systems},  131--139.
\newblock International Foundation for Autonomous Agents and Multiagent
  Systems.

\bibitem[\protect\citeauthoryear{Jim{\'e}nez, Thomas, and
  Torras}{2001}]{jimenez20013d}
Jim{\'e}nez, P.; Thomas, F.; and Torras, C.
\newblock 2001.
\newblock 3d collision detection: a survey.
\newblock {\em Computers \& Graphics} 25(2):269--285.

\bibitem[\protect\citeauthoryear{Kiel, Luther, and
  Dyllong}{2013}]{kiel2013verified}
Kiel, S.; Luther, W.; and Dyllong, E.
\newblock 2013.
\newblock Verified distance computation between non-convex superquadrics using
  hierarchical space decomposition structures.
\newblock {\em Soft Computing} 17(8):1367--1378.

\bibitem[\protect\citeauthoryear{Kockara \bgroup et al\mbox.\egroup
  }{2007}]{kockara2007collision}
Kockara, S.; Halic, T.; Iqbal, K.; Bayrak, C.; and Rowe, R.
\newblock 2007.
\newblock Collision detection: A survey.
\newblock In {\em 2007 IEEE International Conference on Systems, Man and
  Cybernetics},  4046--4051.
\newblock IEEE.

\bibitem[\protect\citeauthoryear{Li \bgroup et al\mbox.\egroup
  }{2019}]{li2019large}
Li, J.; Surynek, P.; Felner, A.; Ma, H.; and Satish, K.~T.
\newblock 2019.
\newblock Multi-agent pathfinding for large agents.
\newblock In {\em AAAI}.

\bibitem[\protect\citeauthoryear{Moore and Wilhelms}{1988}]{triangular}
Moore, M., and Wilhelms, J.
\newblock 1988.
\newblock Collision detection and response for computer animation.
\newblock In {\em ACM Siggraph Computer Graphics}.
\newblock ACM.

\bibitem[\protect\citeauthoryear{Neyman and Pearson}{1933}]{neyman1933testing}
Neyman, J., and Pearson, E.~S.
\newblock 1933.
\newblock The testing of statistical hypotheses in relation to probabilities a
  priori.
\newblock In {\em Mathematical Proceedings of the Cambridge Philosophical
  Society}, volume~29,  492--510.
\newblock Cambridge University Press.

\bibitem[\protect\citeauthoryear{Requicha and Voelcker}{1977}]{CSG}
Requicha, A.~A., and Voelcker, H.~B.
\newblock 1977.
\newblock Constructive solid geometry.

\bibitem[\protect\citeauthoryear{Wagner, Willhalm, and
  Zaroliagis}{2005}]{wagner2005geometric}
Wagner, D.; Willhalm, T.; and Zaroliagis, C.
\newblock 2005.
\newblock Geometric containers for efficient shortest-path computation.
\newblock {\em Journal of Experimental Algorithmics (JEA)} 10:1--3.

\bibitem[\protect\citeauthoryear{Wu}{1991}]{wu}
Wu, X.
\newblock 1991.
\newblock An efficient antialiasing technique.
\newblock In {\em Proceedings of the 18th Annual Conference on Computer
  Graphics and Interactive Techniques}, SIGGRAPH '91,  143--152.
\newblock New York, NY, USA: ACM.

\bibitem[\protect\citeauthoryear{Yakovlev and Andreychuk}{2017}]{AASIPP}
Yakovlev, K., and Andreychuk, A.
\newblock 2017.
\newblock Any-angle pathfinding for multiple agents based on sipp algorithm.
\newblock {\em arXiv preprint arXiv:1703.04159}.

\end{thebibliography}
